\documentclass[11pt,a4paper]{article}
\usepackage[hyperref]{emnlp2018}
\usepackage{times}

\usepackage{latexsym}
\usepackage{graphicx}
\usepackage{multirow}

\usepackage{url}

\aclfinalcopy 

\title{A Preordered RNN Layer Boosts Neural Machine Translation in Low Resource Settings}

\author{Mohaddeseh Bastan \\
  Stony Brook University  \\
  {\tt  mbastan@cs.stonybrook.edu} \\\And
  Shahram Khadivi\thanks{* Shahram Khadivi has contributed to this work when he was with Amirkabir University of Technology.} \\
 Amirkabir University of Technology \\
  {\tt khadivi@aut.ac.ir} \\}

\date{}

\begin{document}
\maketitle
\begin{abstract}
  Neural Machine Translation (NMT) models are strong enough to convey semantic and syntactic information from the source language to the target language. However, these models are suffering from the need for a large amount of data to learn the parameters. As a result, for languages with scarce data, these models are at risk of underperforming. We propose to augment attention based neural network with reordering information to alleviate the lack of data. This augmentation improves the translation quality for both English to Persian and Persian to English by up to 6\% BLEU absolute over the baseline models.
\end{abstract}

\section{Introduction}
\label{sec1}
NMT has recently shown promising results in machine translation 
\citep{wu2016google,luong2015effective,bastan2017neural}. 
In statistical machine translation (SMT), 
the problem is decomposed into sub-models and each individual model is trained separately, while NMT is capable of training an end-to-end model.
For instance, in SMT the reordering model is a feature that is trained separately and is used jointly with other features to improve the translation, while in NMT it is assumed that the model will learn the order of the words and phrases itself.

Sequence-to-sequence NMT models consist of two parts, an encoder to encode the input sequence to the hidden state and a decoder that decodes the hidden state to get the output sequence \cite{cho2014properties,bahdanau2014neural}. The encoder model is a bidirectional RNN, the source sentence is processed once from the beginning to the end and once in parallel from the end to the beginning.
One of the ideas that have not been well-explored in NMT so far is the use of existing reordering models in SMT. We propose to add another layer to the encoder that includes reordering information.  
The intuition behind our proposal comes from the improvement achieved by bidirectional encoder model. If processing the source sentence in both directions help sequence-to-sequence model to learn better representation of the context in hidden states, adding the order of the input words as they are appearing in the output sequence as another layer may also help the model to learn a better representation in both context vectors and hidden states. In this paper we investigate this hypothesis that another layer in the encoder to process a preordered sentence can outperform both encoder architecture with two or three RNN layers. We empirically show in the experiments that adding the reordering information to NMT can improve the translation quality when we are in shortage of data.\\
There are a few attempts to improve the SMT using neural reordering models \cite{cui2015lstm,li2014neural,li2013recursive,aghasadeghi2015monolingually}. In \cite{zhang2017incorporating}, three distortion models been studied to incorporate the word reordering knowledge into NMT. They used reordering information to mainly improve the attention mechanism.\\
In this paper, we are using a soft reordering model to improve the bidirectional attention based NMT. This model consists of two different parts. The first part is creating the soft reordering information using the input and output sequence, the second part is using this information in the attention based NMT.\\
The rest of the paper is as follow, in section 2 a review of sequence-to-sequence NMT is provided, in section 3 the preordered model is proposed, section 4 explains the experiments and results, and section 5 concludes the paper.

\section{Sequence-to-Sequence NMT}

\citet{bahdanau2014neural} proposed a joint translation and alignment model  which can both learn the translation and the alignment between the source and the target sequence. 
In this model the decoder at each time step, finds the maximum probability of the output word $y_i$ given the previous output words $y_1 ,..., y_{i-1}$ and the input sequence $X$ as follow:
\begin{equation}
    p(y_i | y_1 , ... , y_{i-1},X) = softmax(g(y_{i-1} , s_i , c_i))
\end{equation}
Where $X$ is the input sequence, 
$g$ is a nonlinear function, $s_i$ is the hidden state, and $c_i$ the context vector using to predict output $i$.
$s_i$ is the hidden state at the current step which is defined as follow:
\begin{equation}
    s_i = f(s_{i-1} , y_{i-1} , c_{i})
\end{equation}
The notation $c_i$ is the context vector for output word $y_i$. 
The context vector is the weighted sum of the hidden states as follow:
\begin{equation}
    c_i = \sum_{j=1}^{T} \alpha_{ij}h_j
\end{equation}
The weights in this model are normalized outputs of the alignment model which is a feed-forward neural network. It uses $s_{i-1}$ and $h_j$ as input and outputs a score $e_{ij}$. This score is then normalized and used as the weight for computing the context vector as follow:
\begin{equation}
    \alpha_{ij} = \frac{\exp{(e_{ij})}}{\sum_{k=1}^{T} (\exp{e_{ik}})}
\end{equation}

In the encoder, a bidirectional neural network is used to produce the hidden state $h$. For each input word $x_i$ there is a forward and a backward hidden state computed as 
follow respectively:
\begin{equation}
\label{eq:forward}
    \overrightarrow{h_i} = \overrightarrow{f}(\overrightarrow{h_{i-1}},x_i) 
    \end{equation}
    \begin{equation}
    \label{eq:backward}
    \overleftarrow{h_i} = \overleftarrow{f}(\overleftarrow{h_{i-1}},x_i)
\end{equation}

Forward and backward hidden states are then concatenated to produce the hidden state $h_i$ as follow:
\begin{equation}
h_i = [ \overrightarrow{h_i} , \overleftarrow{h_i}]
\end{equation}
\section{Preordered RNN}
The attention-based model is able to address some of the shortcomings of the simple encoder-decoder model for machine translation. It works fine when we have plenty of data. But if we are in lack of data the attention-based model suffers from lack of information for tuning all the parameters. We can use some other information of the input data to inject into the model and get even better results.
In this paper, a model is proposed using reordering information of the data set to address the issue of shortage of data. Adding this information to the model, it can improve the attention-based NMT significantly.


\subsection{Building Soft Reordered Data}
Adding a preordered layer to the encoder of the sequence model boosts the translation quality. This layer add some information to the model which previously hasn't been seen. 
The preordered data is the source sentence which is reordered using the information in target sentence. The reordered models have been used in statistical machine translation and they could improve the translation quality~\cite{visweswariah2011word,tromble2009learning,khalilov2010source,collins2005clause,xia2004improving}. \\
To obtain the soft reordering model, we first need to have the word alignment between the source and the target sentences, then by using heuristic rules we change the alignment to reordering. The reordered sequence model is built upon the alignment model. First by using GIZA++ \cite{och03:asc}  the alignment model between the input sequence and output sequence is derived. The main difference between reordering and alignment is that alignment is a many-to-many relation, while the reordering is a one-to-one relation. It means one word in the input sequence can be aligned to many words in the output sequence while it can be reordered to just one position. The other difference is that the alignment is a relation from input sequence space to output sequence space while the reordering is a relation from input sequence space to itself.  So we propose some heuristic rules to convert the alignment relation to the reordering relation as follow:
\begin{itemize}
 \setlength\itemsep{-0.2em}
    \item If a word $x$ in the input sequence is aligned to one and only one word $y$ in the output sequence, the position of $x$ in the reordering model will be the position of $y$.
    \item If a word $x$ in the input sequence is aligned to a series of words in the output sequence, the position of $x$ in the reordering model will be the position of the middle word in the series\footnote{We arbitrary round down the even number. For example, the middle position between 1,3,5,7 is the 3rd position.}. 
    \item If a word in the input sequence is not aligned to any word in the output sequence, the position for that word is the average positions of the previous and the next word.
    
\end{itemize}
These heuristic rules are inspired by the rules which have been proposed in \cite{devlin2014fast}. The difference is that they are trying to align one and only one input word to all output words, but we are trying to align each word in the input sequence to one and only one position in the same space. \\
The order of applying these rules is important. We should apply the first rule, then the second rule and finally the third rule to all possible words. If a word is aligned to a position but that position is full, we align it to the nearest empty position. We arbitrarily prioritize the left position to the right position whenever they have the same priority. At the end, each word is aligned with only one position, but there may be some positions which are empty. We just remove the empty positions between words to map the sparse output space to the dense input space. We can build the reordered training data using these rules and use them for training the model. In the next section, we see how the reordered data is used in the bidirectional attention based NMT.

\subsection{Three-layer Encoder}
The bidirectional encoder has two different layers. The first layer consists of the forward hidden states built by reading the input sequence from left to right and the second layer consists of the backward hidden states, built by reading the input sequence from right to left. We add another hidden layer to the encoder which is built by reading the input sequence in the reordered order. We build the hidden layer of the reordered input as follow: 

\begin{equation}
    hr_i = f(hr_{i-1},xr_i)
\end{equation}
Here $xr_i$ is the word in position $i$ of the reordered data and $hr_i$ is the hidden representation of $x_i$ in reordered set. The function for computing $hr$ is the same as in equation \ref{eq:forward} and \ref{eq:backward}.
Then the hidden representation $h$  is computed by concatenating the forward hidden layer, backward hidden layer and reordering hidden layer as follow:
\begin{equation}
h_i = [ \overrightarrow{h_i} , \overleftarrow{h_i}, hr_{i}]
\end{equation}

\section{Experiments}

\begin{table}[t!]
\begin{center}
\begin{tabular}{c| c c c }

\multirow{2}{2em}{Corpus} &\multirow{2}{2em}{\#sents}& \multicolumn{2}{c}{\#words} \\ 
&&English&Persian \\ \hline 
 Training  &26142&264235&242154 \\ 

 Development &276&3442&3339 \\ 
 Test  & 250 &2856&2668 \\

\end{tabular}
\end{center}
\caption{\label{table:dataset} The statistics of data set}
\end{table}
The proposed model has been evaluated on English-Persian translation data set. We believe that adding the reordering information results in a better model in case of low resource data. We evaluate the translation quality based on BLEU \cite{papineni2002bleu} and TER \cite{snover2006study}. 
For implementation we use the Theano \cite{bergstra2011theano} framework. 

\subsection{Dataset}

We use  Verbmobil \cite{bakhshaei2010farsi}, an English-Persian data set, this data set can show the effectiveness of the model on scarce data resources. The detailed information of the data set is provided in \ref{table:dataset}. In this table, the number of words, shown with \#words, number of sentences in each corpus is shown in column \#sents. 
\subsection{Baseline}
The baseline model for our experiments is the bidirectional attention based neural network \cite{bahdanau2014neural} as explained in section 1. There are various papers to improve the basic attention based decoder of the baseline, among all we used guided alignment \cite{chen2016guided}.
\subsection{Reordering Development and Test Set}

For building the reordered training set, we use alignment model and heuristic rules. For development and test set, as we don't have access to the target language, we use a preordering algorithm proposed in \cite{nakagawa2015efficient}. This algorithm is the improved version of preordering algorithm based on Bracketing Transduction Grammar (BTG). Briefly, this algorithm builds a tree based on the words, so that each node has a feature vector and a weight vector. Among all possible trees on the data set, the tree with maximum value for the weighted sum of the feature vectors is chosen as reordering tree. Using a projection function, the tree is converted into the reordered output. \\
This algorithm also needs part of speech  (POS) tagger and word class. for Persian POS tagging we use CMU NLP Farsi tool \cite{feely2014cmu} and for the English POS tagging, we use Stanford POS tagger \cite{toutanova2003feature}. For word class we use the GIZA ++ word class which is an output of creating alignment.

\subsection{Results}

\begin{table}[t!]
\begin{center}
\begin{tabular}{c c c c}
\multicolumn{4}{c}{Reordering Method}   \\
\hline

 Training Set  &Dev/Test Set & BLEU & TER  \\\hline
 HG & BI&30.53&53.25  \\ \hline
 BI & BI &27.91& 56.68  \\ \hline
 BG & BG &25.93& 58.1  \\ \hline

\end{tabular}
\end{center}

\caption{\label{table:results} The comparison between different reordering methods on Verbmobil data. HG means the data reordered using alignment model with GDFA and heuristic rules, BI and BG means the data is reordered on intersection alignment and GDFA alignment, respectively, both using \cite{nakagawa2015efficient} algorithm.}
\end{table}

We analyzed our model with different configurations. First we use different methods to reorder training, development and test set. The results are shown in \ref{table:results}. In this table, the best results of different combinations for building reordered data is shown. HG means for building the reordered data, heuristic rules and alignment with GDFA \cite{koehn2005europarl}  is used. BI means the algorithm in \cite{nakagawa2015efficient} and alignment with intersection method is used to build the reordered data, BG means alignment with GDFA and reordering algorithm in \cite{nakagawa2015efficient} is used. The best possible combinations are shown in Table~\ref{table:results}.

In Table~\ref{table:baseline} we can compare the best 3-layer network with two different 2-layer networks. The 3-layer network has apparently three layers in the encoder, the first two layers are the forward and the backward RNNs, the third layer is again an RNN trained either on the reordered source sentence or the original sentence. The 2-layer network refers to the bidirectional attention based NMT as described in Section 2. This model id trained once with the original sentence, and once with the reordered sentence.
As we see, reordering the input can improve the model. It shows that the information we are adding to our model is useful. So using the best 3layer model can use both information of reordering and information of the ordered data, so it can improve the translation model significantly. Also we see that adding just a simple repeated layer to bidirectional encoder, can improve the model. But not as much as the reordered layer. Finally, the ensemble of different models has the best results. \\
There are different interpretations behind this results. Because NMT has too many parameters, it is difficult for scarce data to learn all of the parameters correctly. So adding explicit information using the same data can help the model to learn the parameters better. In addition, although we expect that all the statistical features we use in SMT automatically be trained in NMT, but it can not learn them as well as SMT. 

\begin{table}[t!]
\begin{center}
\begin{tabular}{|l|l|l|l|}
\multicolumn{4}{c}{Reordering Method}   \\ \hline
 Data set & Model & BLEU & TER  \\ \hline

\multirow{5}{*}{ En $\rightarrow$ Pr} & Baseline SMT &30.47 & -- \\ 
 &Baseline NMT & 27.42 & 50.78  \\ 
 &3-layer RpL & 27.58&50.04  \\ 
 &2-layer RI  & 29.6&50.96  \\ 
 &3-layer RL  & 31.03&47.5  \\ 
 &\textbf{Ensemble}&\textbf{32.74}&\textbf{46.4}  \\ \hline

\multirow{5}{*}{Pr $\rightarrow$ En}&Baseline SMT&26.91& -- \\ 
 &Baseline NMT & 26.12 & 55.87  \\ 
 &3-layer RpL & 26.38&57.42  \\ 
 &2-layer RI & 27.52&54.12  \\ 
 &3-layer RL & 30.53&53.25  \\
 &\textbf{Ensemble}&\textbf{32.17}&\textbf{52.12} \\ \hline




\end{tabular}
\end{center}

\caption{\label{table:baseline} The comparison between different models. base line in SMT is the result of translation in statistical machine translation. The base line NMT is the bidirectional attention based neural network using guided alignment \cite{bahdanau2014neural,chen2016guided}. The 2layer RI is the basic model with reordered input. The 3layer RL is the model proposed in this paper. The 3layer RpL is a 3layer model with two forward and one backward layers (No reordering layer). The ensemble model is the combination of different models.}
\vspace{-1em}
\end{table}

\section{Conclusion}
In this paper we analyzed adding reordering information to NMTs. NMTs are strong because they can translate the source language into target without breaking the problem into sub problems. In this paper we proposed a model using explicit information which covers the hidden feature like reordering. The improvements is the result of adding extra information to the model, and helping the neural network learn the parameters in case of scarce data better.

\bibliography{emnlp2018}
\bibliographystyle{acl_natbib}

\end{document}